\documentclass[preprint,5p,times,twocolumn,authoryear]{elsarticle}

\usepackage{amssymb,amsthm}
\usepackage{xcolor}
\usepackage{tabularx}
\usepackage{booktabs}
\usepackage{multirow}
\usepackage{stfloats}

\usepackage[hidelinks]{hyperref}

\usepackage{tikzconfig}
\usepackage{tikz}

\theoremstyle{definition}
\newtheorem{definition}{Definition}
\newtheorem{example}{Example}
\newtheorem*{excont}{Example \continuation}
\newcommand{\continuation}{??}
\newenvironment{continuedexample}[1]
 {\renewcommand{\continuation}{\ref{#1}}\excont[continued]}
 {\endexcont}

\journal{AI Open}

\newcommand{\tuple}[1]{\ensuremath{\langle #1 \rangle}}
\newcommand*\emptycirc[1][1ex]{\tikz\draw (0,0) circle (#1);} 
\newcommand*\fullcirc[1][1ex]{\tikz\fill (0,0) circle (#1);} 

\newcommand{\tocheck}[1]{\textcolor{blue}{#1}}
\renewcommand{\tocheck}[1]{#1}

\begin{document}

\begin{frontmatter}

\title{An Ecosystem for Personal Knowledge Graphs: A Survey and Research Roadmap}

\author{Martin G. Skj{\ae}veland}
\author{Krisztian Balog}
\author{Nolwenn Bernard}
\author{Weronika \L{}ajewska}
\author{Trond Linjordet}

\address{University of Stavanger, Norway}

\begin{abstract}
    This paper presents an ecosystem for personal knowledge graphs (PKG\tocheck{s}), commonly defined as resources of structured information about entities related to an individual, their attributes, and the relations between them. PKGs are a key enabler of secure and sophisticated personal data management and personalized services. However, there are challenges that need to be addressed before PKGs can achieve widespread adoption. One of the fundamental challenges is the very definition of what constitutes a PKG, as there are multiple interpretations of the term. We propose our own definition of a PKG, emphasizing the aspects of (1) data ownership by a single individual and (2) the delivery of personalized services as the primary purpose. We further argue that a holistic view of PKGs is needed to unlock their full potential, and propose a unified framework for PKGs, where the PKG is a part of a larger ecosystem with clear interfaces towards data services and data sources. A comprehensive survey and synthesis of existing work is conducted, with a mapping of the surveyed work into the proposed unified ecosystem. Finally, we identify open challenges and research opportunities for the ecosystem as a whole, as well as for the specific aspects of PKGs, which include population, representation and management, and utilization. 
\end{abstract}


\end{frontmatter}


\section{Introduction}
\label{sec:introduction}

The concept of a \emph{personal knowledge graph} (PKG) broadly refers to ``a resource of structured information about entities personally related to its user, their attributes and the relations between them''~\citep{Balog:2019:ICTIR}.
There are clear advantages to using PKGs for personal data management and storage as well as for supporting personalized services, such as search and recommendation: the user remains in control of their data and can decide what type of access to grant to which service on what part of their PKG.

While the overall vision is appealing, and there is a growing body of work on PKGs\tocheck{~\citep{Tiwari:2023:book}}, several challenges remain before PKGs can deliver on their promise and enjoy more widespread adoption.
One of the remaining open fundamental questions is the very definition of a PKG.
There appear to be multiple interpretations of what constitutes a PKG, and often the term is only implicitly defined.
\tocheck{Often}, the concept is used in the sense of \emph{personalized knowledge graphs} (or \emph{personal interest graph}), i.e., a subset of an existing knowledge graph that characterizes the interests of a given individual.
This interpretation leaves out the most essential feature of a \emph{personal knowledge graph}: having the individual in control of their data (to decide who and what services get read or write access).
As we will illustrate with two specific scenarios below, this distinction is critical for unlocking the full potential of PKGs.
\tocheck{While Solid (Social Linked Data) is a prominent Semantic Web initiative promoting personal data storage in personal online data stores (PODs)~\citep{Sambra:2016:techrep}, its focus primarily lies on data ownership and management. In comparison, our work takes a more holistic approach to the PKG problem space, encompassing several areas where Solid currently falls short, such as seamless integration of various data sources and user-friendly administration.}

The main contributions of this work are threefold.
First, we present a unifying framework for PKGs, emphasizing that they need to form part of a larger ecosystem and interact with other services and data sources to reach their full potential.  We argue that such a holistic view is needed in order to understand what the key features and requirements of PKGs are and how different research areas can contribute to addressing these.  This holistic treatment of PKGs is perhaps the most important difference that distinguishes ours from previous work, which tends to focus on a single aspect or a narrow use of PKGs in a given application context.
Second, we survey and synthesize existing work using our unified framework. Specifically, we organize our discussion around three main aspects of PKGs: (1) population, (2) representation and management, and (3) utilization.
Third, we identify a set of open challenges and outline research opportunities, both for the ecosystem as a whole and specific to each of the three main aspects listed above.

\subsection{Motivating Scenarios}
\label{sec:scenarios}
\label{sec:motivating-scenarios}

To illustrate the use of PKGs, we present two motivating scenarios that will be used as running examples throughout the paper. The proposed scenarios differ significantly in their complexity and challenges involved. The first scenario addresses a straightforward utilization of a PKG for a personalized recommender system that exemplifies how information integrated from varying data sources may be  leveraged for providing highly contextualized services. Whereas, the second scenario presents a more complex usage of a PKG for managing and sharing health information that involves additional challenges, such as handling inconsistencies between different data sources and dealing with sensitive data.

\begin{example}[Personal Trainer Assistant]\label{ex:training}
Representing a family of applications around personalized recommender systems, a \emph{personal trainer assistant} service can suggest a training plan using personal information regarding the physical condition, diet, and previous injuries of the person that are stored in a PKG. The training recommendations can be synchronized with the user's calendar and include different forms of activities depending on individual preferences. 
The PKG may integrate with external sources of personal information, for example, Facebook or YouTube, to access information about attended sport events or subscribed workout channels. External sources of public data such as Wikidata may also be integrated with the PKG to get more in-depth information about diseases, dietary requirements, or physical injuries.
The personal trainer application is not limited to providing personalized recommendations in response to explicitly expressed information needs, but may include proactive suggestions related to other integrated services in response to user's activity.
For example, it might invite the user to a suitable Strava\footnote{\url{https://www.strava.com/}} challenge if it becomes available at a nearby location.
\end{example}

\begin{example}[Sharing Health Information]\label{ex:health}
In addition to forming a basis for personalized recommendations, a PKG may also constitute a tool for managing and sharing personal information. 
For example, in the case of a complex disease, a person may be a patient or client to a variety of different health service providers. In such a situation, it is typically a challenge for different services to stay updated on the current status of the person's disease and treatment (e.g., the current medication regimen and relevant medical facts in  family history).
Current clinical practice relies on documenting many important facts in prose text journals, which in turn are often too voluminous and dense for health personnel to exhaustively read.
A PKG containing medical health information could provide a means for the patient, and their family, to give different health service providers access to pertinent information and thus facilitate appropriate diagnosis and treatment.
In principle, each provider could have a pre-defined expected view or subset of the PKG that is known to be of interest to their technical specialty. For example, when coming to a new dentistry clinic the patient with such a PKG could easily share with their new dentist all the facts generated in treatment with their previous dentist. 
\end{example}

In both scenarios, the PKG acts as a personal data storage that its owner can use to securely store data of different types, including public, private, and sensitive data, and can grant different services customized access to specific parts of this data. The PKG may be set up to access data from other sources and also to synchronize data back to these sources. Overall, the PKG is a key enabler of personalized services that are offered to its owner and to users that the owner has granted access.

\subsection{Outline}

\autoref{sec:related-work} presents existing work from multiple fields related to the concept of PKGs.
In \autoref{sec:pkg} we introduce a terminology for characterizing PKGs, describe the central components that collectively form a PKG ecosystem, and identify the main processes that are relevant for the construction and use of a PKG.
We also contrast and relate our definition of a PKG to other similar concepts.
\autoref{sec:survey} surveys related work by categorizing them according to the terminology and aspects introduced in \autoref{sec:pkg}.
In \autoref{sec:challenges} we identify gaps between the current state of the art in the field and the requirements proposed by the scenarios described in \autoref{sec:scenarios}, and propose directions for future work.
Finally, we conclude in \autoref{sec:conclusion}.

\section{Related Work}
\label{sec:related}
\label{sec:related-work}

In this section, we discuss how personal knowledge graphs are related to knowledge graphs, personalized information access, personal information management, knowledge extraction and knowledge base population, and the Semantic Web.

\subsection{Knowledge Graphs}
\label{sec:related:kg}

The term \emph{knowledge graph} (KG),
perhaps best known due to its popularization 
by Google under the name ``Google Knowledge Graph'' in 2012,\footnote{\url{https://blog.google/products/search/introducing-knowledge-graph-things-not/}}
is defined in many different ways---dated both before and after 2012 (cf.~\citet{Hogan:2021:kg-book}).
We choose to follow the broad definition by \citet{Hogan:2021:kg-book}:
``a knowledge graph is a graph of data intended to accumulate and
convey knowledge of the real world, whose nodes represent entities of
interest and whose edges represent relations between these
entities. The graph of data (a.k.a. data graph) conforms to a graph-based
data model, which may be a directed edge-labelled graph, a property
graph, etc.''

\tocheck{In the literature, the terms \emph{knowledge base} and \emph{knowledge graph} are often used interchangeably.} 
We consider the term \emph{knowledge base} (KB) to be a generalization of the term knowledge graph and take this to mean a knowledge graph that does not necessarily conform to the graph-based data model.

One of the benefits of using a graph model for representing knowledge is that graph data may easily be extended and integrated with other graph datasets
in a dynamic and incremental manner
without the need to conform to a particular predefined schema. Additionally, it supports representing entities for which some relationships or data values are missing or not known.
The versatility of the graph model also allows for representing many types of data formats and kinds of data as long as it can be encoded into nodes and edges. This allows for example schema information to be represented alongside the data, which makes it possible to manage schema information using the same tools and methods as for the graph data.

Knowledge graphs are a natural choice to use for PKGs to represent personal knowledge since such data, depending on the use case, can be
highly dynamic and disparate since it is
subject to frequent changes and updates;
express facts of different granularity, accuracy and modality;
and originate from a wide range of sources that may use different representation languages and formats.

\subsection{Personalized Information Access}
\label{sec:related:pia}

The amount of information on the Web has made it difficult and time-consuming for a user to search for specific information. 
The gap between how well a system could perform if it tailors results to the individual, and how well it performs by returning results designed to satisfy everyone is referred to as the \emph{potential for personalization} \citep{Teevan:2010:ACM}. It represents the potential improvement of the returned ranking to be achieved by targeting the needs of a specific individual.

Based on this idea, 
personalized services have been developed 
to tailor the information presented to a user based on the representation of their long-term interests~\citep{White:2016:Book}.
Previous studies have shown that using information such as document content or concepts, browser history, query history, and user groups enhances the ranking of documents retrieved for individual users~\citep{Matthijs:2011:WSDM,Sontag:2012:WSDM,Teevan:2009:WSDM,Dou:2007:WWW,Shen:2005:CIKM}. Collecting such information over a significant period of time allows for the creation of a user model that represents the user's long-term interests.

There are two main approaches to provide personalized search results based on the representation of a user's interests.
On one hand, the top-\emph{n} retrieved documents are re-ranked based on the user model in order to promote documents matching user's preference to the top~\citep{Sontag:2012:WSDM,Teevan:2005:SIGIR}, hence, increasing their chance to be inspected. 
On the other hand, the user model can be considered as a part of the ranking algorithm, i.e., the retrieved documents for a query are biased towards the user's preferences~\citep{Agichtein:2006:SIGIR}.

However, personalization is not an optimal solution in dynamic, biased, and data-intensive environments where the user's information needs are changing constantly. The epistemic bubbles are an example of this; these bubbles exclude relevant information, perhaps accidentally, because it does not match the user's interests~\citep{Nguyen:2020:Episteme}. For example, two users who have different political positions will not likely see the same information, as personalization will filter out the information that contradicts the user's beliefs.

The question of user privacy is a recurrent issue in the field of personalized information access. Indeed, there is a tension between privacy and personalization, as the latter needs to collect user information that can reveal private information, such as political inclination and profession~\citep{Shen:2007:SIGIRForum}. Therefore, a compromise between how much the users agree to share and how much user information is needed to provide a personalized service should be considered~\citep{Panjwani:2013:CHI}.
Our proposed PKG ecosystem places a strong emphasis on privacy. Indeed, the owner of the PKG has a more direct control over the data they share with each service, allowing them to decide how to balance the trade-off between privacy and personalization.

\subsection{Personal Information Management}
\label{sec:related:pim}

Personal information management (PIM) ``refers to the practice and the study of the activities a person performs in order to acquire or create, store, organize, maintain, retrieve, use, and distribute information in each of its many forms [...] as needed to meet life's many goals [...] and to fulfill life's many roles and responsibilities''~\citep{Jones:2017:Bookchapter}.
While the origins of PIM may be traced back as far as the seminal article ``As We May Think'' by \citet{Bush:1945:AM}, where he describes the concept of the memex that would make knowledge more accessible, the inception of contemporary PIM dates back to the formation of a special interest group at the 2004 CHI Conference on Human Factors in Computing Systems~\citep{Bergman:2004:CHI}, followed by an NSF workshop in 2005~\citep{Jones:2005:PIM}.
PIM places special emphasis on the organization and maintenance of personal information items for later use and repeated reuse~\citep{Jones:2017:Bookchapter}.
An \emph{information item} is defined to be an encapsulation of information in a persistent form that can be managed (i.e., created, stored, copied, moved, deleted)~\citep{Jones:2017:Bookchapter}.
Examples of information items include files, emails, web pages, posts, and status updates on social media platforms.
Importantly, information items are not restricted to digital forms, but can also be paper-based documents.
There are several ways in which information can be \emph{personal}: it can be 
(P1) controlled/owned by the individual (e.g., email messages in one's account, files on a hard drive or in a cloud service),
(P2) about the individual (e.g., credit history, medical records, web search history),
(P3) directed towards the individual (e.g., emails, web ads, tweet mentions),
(P4) sent/posted/shared by the individual (e.g., sent emails, published articles),
(P5) things experienced by the individual (e.g., web history, photos, videos),
(P6) potentially relevant/useful to the individual (e.g., future job, home, partner)~\citep{Jones:2017:Bookchapter}.
As noted by \citet{Jones:2017:Bookchapter} ``the senses in which information can be personal are not mutually exclusive.'' For example photos taken at a given event may be owned (P1), about (P2), shared (P4), and experienced (P5) by the same person.

PIM is associated with three main activities~\citep{Jones:2017:Bookchapter}: 
(1) \emph{keeping activities} include decisions concerning what subset of the encountered information and how should be kept for later use,
(2) \emph{finding/re-finding activities} include explicit searches as well as various navigation activities performed to locate information,
(3) \emph{meta-level activities} focus on connecting information with needs and involve organization, maintenance, and making sense and use of personal information.
A distinctive characteristic of PIM research is its strong focus on the human perspective: identifying patterns of behavior in how people approach different forms of information using various computer-based tools.
For example, recurrent themes of discussion include the use of folders versus tags~\citep{Bergman:2013:JASIST,Civan:2008:ASIST,Voit:2012:CHI} and navigation versus search~\citep{Bergman:2008:TOIS,Bergman:2013:PUC,Fitchett:2005:IJHCS,Teevan:2004:CHI}.
Many excellent studies focus on how people use and organize specific forms of information, e.g., their email~\citep{Bellotti:2003:CHI,Capra:2013:JASIST,Hanrahan:2015:CHI,Whittaker:2011:CHI} 
and bookmarks~\citep{Abrams:1998:CHI,Boardman:2004:CHI,Jones:2002:ASIST}, but this also makes the field of PIM fragmented.
PIM is also related to the notion of \emph{quantified self}, which is concerned with the tracking of personal activities, often through a dedicated hardware device (e.g., physical fitness monitors and activity trackers such as smartwatches)~\citep{Gurrin:2014:FnTIR}. 

PIM is closely related to the notion of PKGs, but there are several key differences:
\begin{itemize}
    \item PIM has a strong emphasis on human activities around managing personal information, i.e., the human-computer interaction is in focus. PKGs center around the information itself, how it can be represented and utilized across services and applications.
    \item The atomic units in PIM are information items, while PKGs operate on facts, which is a finer granularity.  Also, PKGs assume digital information, while PIM in the broader sense also includes paper-based documents.
    \item Underlying all PIM activity types is an \emph{implicit} effort to make sense of the available information.  In PKGs, sense-making is \emph{explicit} in the representation of information as facts.    
    \item All PIM activities are driven by the aim to assist in satisfying the user's \emph{information needs}. PKGs have a broader scope. For instance, in Example~\ref{ex:training}, the personal assistant can proactively take initiative, without addressing any existing information need. Another use of PKGs is to share data with others, as illustrated with the case of health service providers in Example~\ref{ex:health}.
    \item Integration is part of both PIM and PKGs, but for different reasons: in PKGs it allows for providing better services, while in case PIM it helps to counter information fragmentation.
\end{itemize}
PKGs can be leveraged in PIM for organizing personal information in a finer granularity (i.e., as facts as opposed to files/documents).  In our envisaged PKG ecosystem, privacy and access management would be taken care of, and PIM could focus on building personalized services that utilize or help maintain this information.

\subsection{Knowledge Extraction and Knowledge Base Population}
\label{sec:related:kbp}

A wealth of information resides in unstructured or semi-structured format (text documents, social media posts, multimedia files, etc.) that is not readily available in knowledge bases.
Knowledge bases can be augmented by extracting structured information from these sources. 
\emph{Knowledge acquisition} (a.k.a. \emph{knowledge harvesting}) refers to the process of extracting information on entities and relationships from a large data corpus (e.g., the Web) and turning them into machine-readable facts~\citep{Weikum:2020:FnTDB}. 

\emph{Knowledge base population} is a more specific task within the broader problem space of knowledge acquisition, focusing on the augmentation of an existing KB with entities, types, attributes of entities, and relationships between entities~\citep{Ji:2011:HLT}. 
A key component of KB population is \emph{entity linking}, which is concerned with the resolution of entity mentions detected in text to unique identifiers in a KB; it is typically performed by leveraging contextual information and existing knowledge bases~\citep{Balog:2018:Book}.
It is important for a KB to have a clean and expressive taxonomy of types (classes) and that these are populated with uniquely identified entities.
\emph{Class-instance acquisition} aims at obtaining additional entities that belong to a given entity type (a.k.a. class)~\citep{Pantel:2009:EMNLP} and obtaining additional entity types for a given entity~\citep{Gangemi:2012:ISWC}.
The process of populating the KB with new facts about entities, i.e., additional attributes (with literal values) and relationships (with other entities) is often referred to as \emph{slot-filling}~\citep{Ji:2011:HLT}.
Extraction is traditionally approached using pattern- and rule-based techniques. Restricted kinds of patterns may be learned automatically from examples~\citep{Sarawagi:2008:FnTDB}.
More recently, the task is viewed either as a classification or as a sequence-tagging problem~\citep{Weikum:2020:FnTDB}.

What we have discussed above, populating an existing KB with additional facts about entities, is an instance of \emph{targeted} and \emph{closed} information extraction.
Targeted, because it focuses on the extraction of a predefined set of predicates (attributes and relationships) for specific entities, and closed in a sense that all entities, types, and relationships already have canonicalized (unique) identifiers~\citep{Balog:2018:Book}.
Open information extraction addresses the discovery of new attributes and relationships and their organization into a canonicalized format with clean type signatures~\citep{Weikum:2020:FnTDB}.
\emph{Novel entity detection} deals with the discovery of out-of-KB entities; it involves identifying and classifying previously unseen or unknown entities within a given text~\citep{Lin:2012:EMNLP}.
\emph{Class-attribute acquisition} refers to the discovery of relevant attributes (and types of relationship) of classes~\citep{Pasca:2007:IJCAI}.
\emph{Predicate discovery} is the task of extracting predicate-argument structures, where the predicate and two or more arguments take the form of short natural language phrases extracted from an input sentence~\citep{Weikum:2020:FnTDB}.
Traditionally, most open information extraction methods relied on patterns and rules~\citep{Fader:2011:EMNLP}.
More recently, neural approaches have been proposed, following the success of deep learning models on various NLP tasks~\citep{Cui:2018:ACL}.
However, canonicalization of the resulting facts for inclusion in a KB, especially of the newly seen phrases describing predicates, remains an open challenge~\citep{Weikum:2020:FnTDB}.

\subsection{The Semantic Web and Semantic Web Technologies}

\emph{The Semantic Web}~\citep{Berners-Lee:2001:SA} is an extension of the World Wide Web (the Web) that combines standards for knowledge representation with established web standards and architecture with the aim of making data available on the Web machine readable and ``understandable.'' This is done by using formal languages with defined syntax and semantics, such as the Resource Description Framework (RDF)\footnote{\url{http://www.w3.org/TR/rdf11-concepts/}} and the Web Ontology Language (OWL),\footnote{\url{http://www.w3.org/TR/owl-overview}} to encode the data and its semantics. The original vision of the Semantic Web is expressed by Tim Berners-Lee as one where computers
``become capable of analyzing all the data on the Web---the content, links, and transactions between people and computers. A `Semantic Web', which makes this possible, has yet to emerge, but when it does, the day-to-day mechanisms of trade, bureaucracy and our daily lives will be handled by machines talking to machines.'' In the seminal paper ``The Semantic Web''~\citep{Berners-Lee:2001:SA}, this is illustrated by an example reminiscent of our motivating scenarios: in the example, Pete and Lucy's (personal) \emph{Semantic Web agents} together organize and schedule a series of health treatments for their mom so that they fit her insurance plan, minimize traveling time, and fit with their own schedules. The agents are able to read and understand the prescribed treatment from their mother's doctor, listings of specialist providers, map data and personal calendar data, and to communicate with each other autonomously and with their users.
It is clear that the vision of The Semantic Web greatly overlaps with the expected outcomes of using PKGs. However, their goals differ in that that personal aspect of PKG is its very definition, while The Semantic Web has the much broader scope of enabling more sophisticated machine to machine communication based on the explicit representation of the semantics of data.

The Semantic Web community has developed a suite of technologies that are relevant for PKGs.
The Resource Description Framework (RDF)\footnote{\url{http://www.w3.org/TR/rdf11-concepts/}} is one of the most important technologies in the \emph{semantic web stack}\footnote{\url{http://www.w3.org/2000/Talks/1206-xml2k-tbl/}} and is a prominent example of a directed edge-labeled graph model \citep{Hogan:2021:kg-book}.
An \emph{RDF graph} is a set of \emph{RDF triples} $\tuple{s, p, o}$ where the
\emph{predicate} $p$ is the labeled edge element that connects
the \emph{subject} element $s$ to the \emph{object} element $o$.
Sometimes in the literature, the term \emph{SPO-triples} is used to designate RDF triples.
The elements of a triple are, with some restrictions, either an \emph{IRI} (Internationalized Resource Identifier), a \emph{literal}, or a \emph{blank node}.
Well-known KGs that use RDF are DBpedia~\citep{Auer:2007:ISWC} and Wikidata~\citep{Vrandevcic:2014:ComACM}.
RDF KGs are often published following principles known as Linked (Open) Data~\citep{Heath:2011:Book} that promotes sharing and connecting data to support machine-readability over the Web using established Web standards such as HTTP (Hypertext Transfer Protocol), IRIs, and RDF. Additionally, RDF KGs are often made available via SPARQL endpoints through which the data may be accessed and queried using the query language and protocol SPARQL.\footnote{\url{http://www.w3.org/TR/sparql11-overview/}}
The Web Ontology Language (OWL) is the de facto standard for representing logical ontologies and is used to define the vocabulary and semantics of the schema used by the data.

\section{Personal Knowledge Graphs}
\label{sec:pkg}

\begin{table*}[t]
    \centering
    \caption{Different interpretations of personal knowledge graphs.}
    \label{tab:interpretation}
    \small
    \begin{tabularx}{\textwidth}{p{1.6cm}|X|X|X}
    \toprule
         & \textbf{PKG (this paper)} & \textbf{PKG \citep{Balog:2019:ICTIR}} & \textbf{Personalized Knowledge Graph} \\
         & (cf. Definition~\ref{def:pkg}) & (cf. Definition~\ref{def:pkg_balog_kenter}) & (cf. Definition~\ref{def:personalized_kg}) \\
    \midrule
        \textbf{Ownership} 
            & Created and maintained by an individual 
            & Created and maintained by an individual 
            & Created and maintained by a service 
            \\ 
        \hline
        \textbf{Public facts}
            & Can incorporate facts from public knowledge graph 
            & Public facts are not explicitly stored, but can be linked 
            & Built with facts from a public/proprietary knowledge graph 
            \\ 
        \hline
        \textbf{Private facts}
            & The owner of the PKG can add private facts (e.g., beliefs) as long as they have the correct format 
            & The owner can add private facts (e.g., beliefs) as long as they are connected to it 
            & Facts that are not of public knowledge cannot be stored (e.g., an individual medication regimen) 
            \\ 
        \hline
        \textbf{Graph structure} 
            & Facts do not need to be connected to the user
            & All facts in the PKG are connected to the user resulting in a spiderweb layout 
            & Facts do not need to be connected to the user
        \\ 
    \bottomrule
    \end{tabularx}
\end{table*}

This section presents our definition of a PKG and discusses how it differs from existing ones (Section~\ref{sec:pkg:def}). Subsequently, the ecosystem within which the PKG exists is introduced and its main aspects are presented (Section~\ref{sec:pkg:ecosystem}).

\subsection{What is a Personal Knowledge Graph?}
\label{sec:pkg:def}

In prior work, the concept of a PKG is often only implied rather than explicitly stated, leaving room for different interpretations. Below, we present the definitions for the various interpretations that exist, starting with our proposed definition.
The key differences between the various interpretations are summarized in Table~\ref{tab:interpretation}.

\begin{definition}[Personal Knowledge Graph] \label{def:pkg}
A \emph{personal knowledge graph} (PKG) is a knowledge graph (KG) where a single individual,
called the owner of the PKG,
has (1) full read and write access to the KG, 
and
(2) the exclusive right to grant others read and write access to any specified part of the KG.
The primary purpose of the PKG is to support the delivery of services that are customized particularly to its owner.
\end{definition}
Note that we do not pose any requirements to the contents of the PKG; the discriminating factor from regular KGs is the administrative rights to the KG that acts as a PKG.
This is different from the definition by \citet{Balog:2019:ICTIR}\tocheck{, which has enjoyed a wide adoption within the research community~\citep{Tiwari:2023:book}}:
\begin{definition}[Personal Knowledge Graph \citep{Balog:2019:ICTIR}]\label{def:pkg_balog_kenter}
A personal knowledge graph is a source of structured knowledge about entities and the relations between them, where the entities and the relations between them are of \emph{personal}, rather than general, importance. The graph has a particular ``spiderweb'' layout, where every node in the graph is connected to one central node: the user.
\end{definition}
Rather than requiring that the owner is explicitly represented in the PKG and that all facts in the PKG are connected to the owner, our definition establishes this relation though the administrative rights to the PKG; as the facts in the PKG can only exist in the PKG by the owner's discretion, they are by definition also personal. \\

\noindent
In the literature, we observe that a majority of the applications focus on the personalization of a service. For example, \citet{Lu:2019:NAACL} and \citet{Zhang:2018:ACL} propose solutions for a dialogue agent to provide personalized answers to a user. To address this problem, it is common to create a representation of a user such as a user profile.
This representation can take the form of a KG which is sometimes referred as a PKG. However, we argue that they actually refer to a \emph{personalized knowledge graph} (a.k.a. \emph{personal interest graph}). \citet{Rastogi:2020:arXiv} distinguish between personalized KGs and PKGs at the level of stored information, with personalized KGs limited to the entities described in the general KGs and PKGs complementing general KGs with additional, personal information about the user.
\begin{definition}[Personalized Knowledge Graph]\label{def:personalized_kg}
A \emph{personalized knowledge graph} is a subset of an existing knowledge graph, restricted to entities and relationships that can characterize the interests of a given individual. 
\end{definition}
In the case of a personalized knowledge graph, the user rarely knows how it is created and with which facts it is populated. Thus, it does not fulfill the ownership criterion of our definition.

\subsection{The PKG Ecosystem}
\label{sec:pkg:ecosystem}

\begin{figure*}
  \centering
  \includegraphics{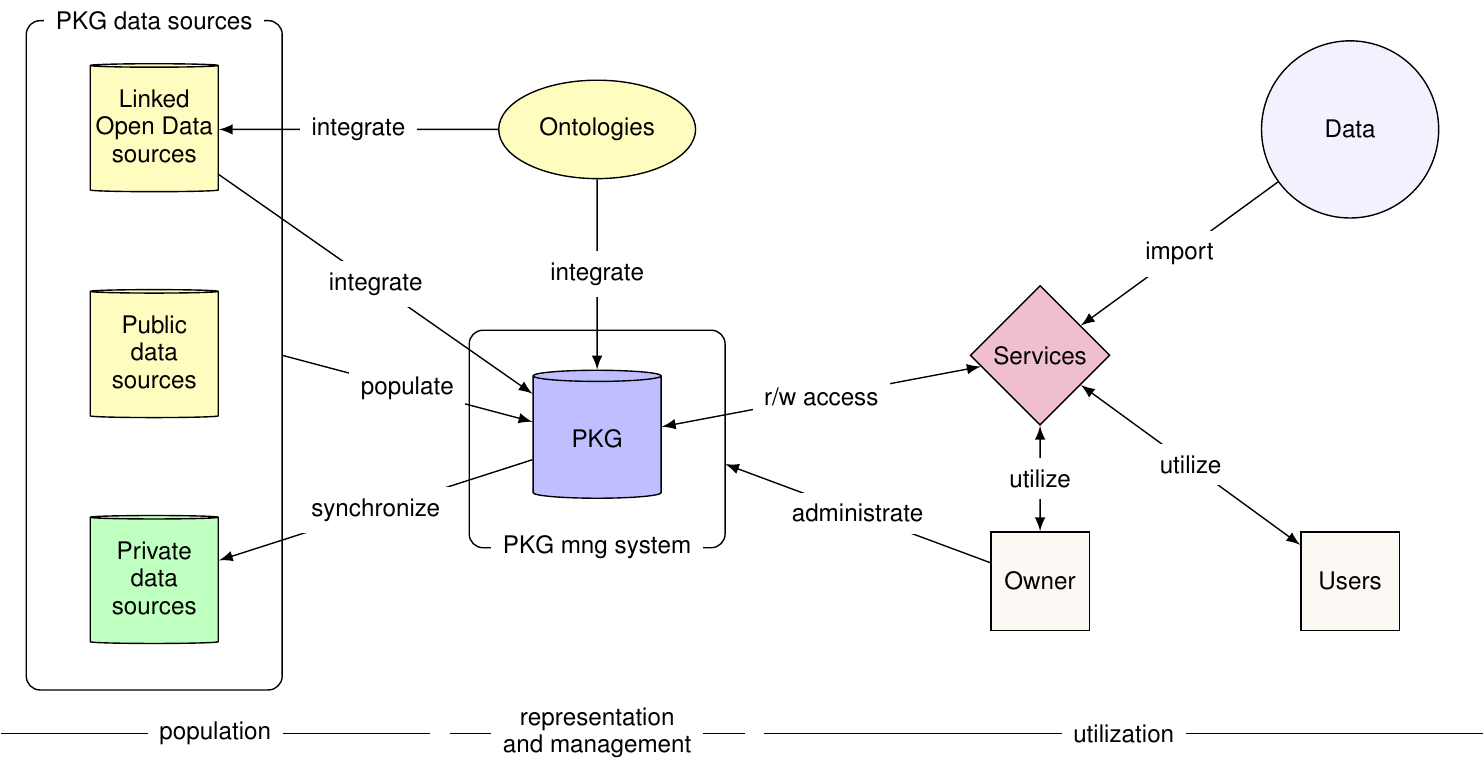}
\caption{PKG ecosystem.}
\label{fig:pkg-arch}
\end{figure*}

As one of the main contributions of this work, we emphasize the necessity of situating the PKG within its broader operational ecosystem in order to conduct meaningful research.
Our proposed unifying architecture for PKG ecosystems is presented in Figure~\ref{fig:pkg-arch}.
We identify three main aspects: \emph{population}, \emph{representation and management}, and \emph{utilization}, all of which are required for the successful engineering and operation of a PKG.

\subsubsection{Population}

Population concerns the aspect of adding data to the PKG from existing data sources or services (cf. Sections~\ref{sec:related:pim} and \ref{sec:related:kbp}). 
We consider three different types of data sources:
\begin{itemize}
    \item \emph{Private data sources} refer to data in any format that are private to the PKG owner, meaning that the owner typically is the only one with the access to read and write to the contents of these sources.
    \item \emph{Public data sources} refer to data in any format that are publicly available.
    \item A specific type of public data sources in the PKG ecosystem are public KGs, often called \emph{Linked Open Data sources}~\citep{Heath:2011:Book}. These data sources are assumed to be in a format that is already compatible with the format of the PKG.  That is, they may be integrated with the PKG without any additional preprocessing or alignment steps, which would typically be required when populating data from the two other types of data sources. 
    Rather, the PKG and Linked Open Data sources both link to vocabulary definitions in external ontologies.
\end{itemize}
The population aspect also includes the process of synchronizing any modifications made to the extracted data in the PKG back to its original data source. This is only relevant for private data sources as these are the only type of sources we assume the owner has access to update.

\begin{continuedexample}{ex:training}
In the Personal Trainer Assistant example,
the PKG is populated with  
private calendar data from the PKG owner's calendar,
data about the owner's sport interests from Facebook and YouTube,
and public health related Linked Open Data from Wikidata.
The relevant data from these sources must be extracted, represented and formatted to fit with the representation requirements of the PKG.
\end{continuedexample}

The population aspect of the PKG ecosystem interfaces with the representation and management aspect, and it is important that the data output from the population processes respect the requirements set by the representation aspect.

\subsubsection{Representation and Management}

This aspect is concerned with \emph{representation}, i.e., 
the logical representation, format and expressivity of the facts and statements that make out the contents of the PKG,
and \emph{management}, i.e., the
organization, storage, retrieval, and access control of the contents of the PKG, whose functionality is collectively made available by the \emph{management system} of the PKG.

The PKG contents should be organized \tocheck{to enable efficient and accurate data capture, access, and update.} 
A PKG should be able to store and organize a wide range of facts or statements, e.g., objective facts (``the capital of Norway is Oslo''), records of personal events (``I visited the dentist May 1st 2020''), logging (``I accessed the medical records in my PKG on May 2nd, 2020''), but arguably also more complex statements such as personal beliefs and probabilities (``I believe the Earth is flat'') and other person's beliefs (``My mom thinks my dentist is very polite'').
To support this range of types of statements, a rich set of metadata and context data becomes vital: where, when, and how did the statement come about, and who holds the fact to be true and to which degree.
This will allow the services that access the PKG data to better assess, select, and use relevant data.
Provenance data is also vital for enabling synchronization with private data sources.
It is reasonable to think that the size of the metadata will greatly exceed the size of the ``actual data.''

Furthermore, the data in the PKG should be semantically rich so that services that access the PKG can exploit the semantic descriptions, metadata, and context descriptions in order to combine PKG data in new ways and to saturate the PKG data with new knowledge
that will improve the data foundation for the services.
This requires the PKG to integrate with several ontologies that can capture the semantics of the  facts stored in the PKG.

\begin{continuedexample}{ex:training}
  In order to efficiently and correctly manage the data collected for the Personal Trainer Assistant service, the PKG management system will benefit from a detailed and accurate set of provenance data.
  For instance, when suggesting training exercises, a preference explicitly set by the owner should carry more weight than a preference inferred from the owner's YouTube viewing history. Additionally, recent data records should take priority over older records.
  By connecting the owner's training preferences to rich semantic descriptions in ontologies and KGs, e.g., ontologies about the human anatomy or KGs about exercises and their benefits, the collected PKG data can be saturated with more data inferred from reasoning over the combination of the collected data and the integrated ontologies. 
  For example, a service could infer by reasoning over ontologies of the human anatomy and exercise injuries that the current runner's knee problem suggests avoiding sport activities similar to running.
\end{continuedexample}

\begin{continuedexample}{ex:health}
In our Sharing Health Information scenario,
assuming that the PKG contains the complete medical history of the owner, a PKG service could monitor the drug prescriptions given to the owner, and by reasoning over drug and medicine ontologies discover and flag new prescriptions that are incompatible with previously given prescriptions or with any of the owner's recorded illnesses and injuries.
\end{continuedexample}

The PKG management system is responsible for
handling the storage and retrieval of data contained in the PKG,
managing and controlling access to and the security of the PKG,
and providing the administrative management service to the owner of the PKG.
Using the management system, the owner should be able to maintain the contents of the PKG,
and grant read and write access to the PKG for other services and users.

\begin{continuedexample}{ex:health}
The PKG management system plays an important role in organizing access for different users to different parts of the PKG data: the owner should have read and write access to the complete contents of the PKG, while the owner's dentist should only have read access for the data content that is relevant for the owner's treatment.
\end{continuedexample}

Note that by definition the owner has read and write access to the PKG; manipulating certain type of information, such as medical records, can have serious negative consequences. We come back to this particular issue later, in Section~\ref{sec:challenges}.

\subsubsection{Utilization}

Utilization is the task of exploiting the PKG data to deliver successful personalized services to its owner and users, and this is where the real value of the PKG is realized.
The services can be personalized only because they access the PKG, but they may also interface with other data sources---in the widest sense, e.g., weather data or Twitter feeds---to deliver value to their users. These services must be able to correctly interface with the PKG management system and understand and exploit the potentially very rich structure of the PKG data. 

\begin{continuedexample}{ex:training}
  The Personal Trainer Assistant service could take the form of a mobile phone application.
  The service would connect to and communicate with the owner's PKG to access the relevant data.
  The PKG management system would control and manage this access, and handle any queries or other requests issued to the PKG by the Personal Trainer Assistant service, such as requests for reasoning over the data with respect to specific ontologies or requests for access to more of the PKG's data.
  The service can also communicate with other data sources than the PKG, e.g., news channels or Twitter feeds, including data that the vendor of the service application make\tocheck{s} available for this particular use, such as news about training trends or sport results, or personalized suggestions made by the service vendor's personnel. The service could then use this data and match it with the relevant data in the PKG to make customized suggestions to the owner.
  The results would then need to be presented to the owner in the application in a user friendly manner.
\end{continuedexample}

\section{Survey}
\label{sec:survey}

In this section, we present previous work on personal knowledge graphs by mapping it to our PKG ecosystem (shown in Fig.~\ref{fig:pkg-arch}). For each work, we identify the aspects and processes that are studied; Table~\ref{tab:survey} presents a summary of our findings, evaluating how well each surveyed work addressed each of the processes identified in our PKG ecosystem framework.\footnote{To illustrate how previous works are summarized in Table~\ref{tab:survey}, consider the ``populate'' column. If a work does not address the ``populate'' process, then that cell of the table is left blank. If the ``populate'' process is acknowledged as a part of the PKG ecosystem, e.g., the importance of extracting facts from data, but no implementation details are mentioned, then that cell of the table is marked with  $\emptycirc$. Finally, if the work addresses the process in practical terms, e.g., as many of the surveyed works research the actual process of extracting facts from data, then that cell in the table is marked with $\fullcirc$.} 
Below, we organize the discussion according to the main aspects, i.e.,  population, representation and management, and utilization, in respective Sections~\ref{sec:survey:population}--\ref{sec:survey:utilization}.
Within each section, studies are presented in chronological order.
Where applicable, we describe the surveyed works with the same terms as in our PKG ecosystem, while also respecting each work's terminology where something more specific or nuanced is being expressed.

\begin{table*}[t]
    \centering
    \caption{Paper categorization based on PKG ecosystem; \fullcirc$\,$ denotes well described process
    , \emptycirc$\,$ denotes briefly described process
    .}
    \label{tab:survey}
    \begin{tabular}{l|cc|ccc|cc}
         & \multicolumn{2}{c|}{\textbf{Population}} & \multicolumn{3}{c|}{\multirow{2}{*}{\parbox{3.5cm}{\centering \textbf{Representation \\and Management}}}
         }
          & \multicolumn{2}{c}{\textbf{Utilization}} \\
         & \rotatebox{70}{Populate} & \rotatebox{70}{Synchronize} & \rotatebox{70}{Represent} & \rotatebox{70}{Integrate} & \rotatebox{70}{\multirow{2}{*}{\parbox{2.5cm}{Manage}}} & \rotatebox{70}{Administrate} & \rotatebox{70}{Utilize} \\ \hline
         \citet{Groza:2007:ISemantics}  & \fullcirc & \emptycirc & \fullcirc & \fullcirc & \fullcirc & \fullcirc& \fullcirc \\ \hline
         \citet{Sambra:2016:techrep} & \emptycirc & \fullcirc & \fullcirc &  &  \fullcirc & \fullcirc & \fullcirc \\ \hline
         \citet{Gyrard:2018:ISWC} & \fullcirc & & \emptycirc  & \emptycirc& & & \emptycirc \\ \hline
        \citet{Mazare:2018:EMNLP} &  & & \emptycirc & & & & \emptycirc \\ \hline
         \citet{Zhang:2018:ACL} &  & & \emptycirc &  & & & \emptycirc \\ \hline
        \citet{Lu:2019:NAACL} &  & & \emptycirc &  & & & \emptycirc \\ \hline
        \citet{Luo:2019:AAAI} & \fullcirc & & \fullcirc & & & & \fullcirc \\ \hline
        \citet{Yen:2019:SIGIR} & \fullcirc & & \fullcirc & \fullcirc & & & \emptycirc \\ \hline
        \citet{Tigunova:2019:WWW} & \fullcirc & & \emptycirc &  & & & \emptycirc \\ \hline
        \citet{Gerritse:2020:ICTIR} & & & & & & & \emptycirc \\ \hline
        \citet{Rastogi:2020:arXiv} & \emptycirc & & \emptycirc & \emptycirc & & & \emptycirc \\ \hline
         \citet{Tigunova:2020:EMNLP}  & \fullcirc & & \emptycirc & & & & \emptycirc \\ \hline
         \citet{Ammar:2021:JMIR} & \fullcirc & & \fullcirc & \emptycirc &   \fullcirc & \fullcirc & \fullcirc \\ \hline
         \citet{Vannur:2021:PAKDD} & \fullcirc & & \emptycirc & \fullcirc & & &  \emptycirc\\ \hline
         \citet{Seneviratne:2021:arXiv} & \fullcirc & & \fullcirc &  & & & \fullcirc \\ \hline
         \citet{Chakraborty:2022:WWW} & \fullcirc & & \fullcirc & & \fullcirc & \emptycirc & \emptycirc \\ \hline
    \end{tabular}
\end{table*}

\subsection{Population}
\label{sec:survey:population}

We first consider previous work in terms of how external data is used to populate a PKG, and how any update in the PKG is synchronized back to the external data sources.
The facts used to populate a PKG can come from both public (e.g., food calorie charts, public healthcare policies) and private (e.g., emails, medical records) data sources.

\citet{Groza:2007:ISemantics} present the NEPOMUK project, an architecture of a social semantic desktop, which is a tool to share data between a local user and other users or applications, inspired by semantic web technologies. 
In NEPOMUK, the PKG, which is only implicitly defined, can be manually populated with RDF data. In addition to manually inserting data, services such as Data Wrapper and Text Analysis are intended to help populate the PKG with application data (email, calendar) and unstructured data (e.g., free text in documents).
The synchronization is enabled only for the resources (e.g., documents) shared with collaborators within one closed group and it does not include propagating PKG changes to the private data sources. However, the described scenario with notifications can be a first step toward the implementation of the synchronization process. 

The Solid (Socially Linked Data) project introduced by~\citet{Sambra:2016:techrep} is a protocol based on Semantic Web technologies with the purpose of re-decentralizing the Web. Solid allows users to own their personal online data stores (PODs), which is similar to our notion of a PKG, and control applications' access to operate on these PODs. Solid does not offer an automatic population of a POD, in particular not from unstructured sources. Synchronization is enabled via the implemented PubSub system based on WebSocket,\footnote{\url{http://www.w3.org/TR/websockets/}} which allows the POD to send live updates and notifications regarding resources of interest.

\citet{Gyrard:2018:ISWC} present a personal health knowledge graph (PHKG) that represents medical and personal data to support the development of a personal health coach or digital health advisor. The PHKG is populated with data from the kHealth project datasets and Linked Open Data resources. Additionally, the Kno.e.sis Alchemy API\footnote{\url{http://wiki.knoesis.org/index.php/Knoesis_Alchemy_of_Healthcare}} is used to extract structured data from different sources such as clinical trials.

\citet{Luo:2019:AAAI} investigate goal-oriented dialog systems which harness models of both (1) the personality and language style of the user, and (2) the preferences of a user with respect to the system's underlying knowledge graph. The proposed dialog system profiles the personality of the user by adding personal information terms, such as gender, age, and diet, for each dialog turn in a slot-filling manner.

\citet{Yen:2019:SIGIR} study the extraction of life events from life logs such as tweets to populate a personal KB for memory recall and life support assistance applications. The population of personal KB is not limited to life events and includes entities from DBpedia~\citep{Auer:2007:ISWC} and Freebase~\citep{Bollacker:2008:SIGMOD}.

The methods for inferring user attributes from a sequence of utterances presented \tocheck{by}~\citet{Tigunova:2019:WWW} are used to create a personal KB that can serve as a distant source of knowledge for personalization. The authors propose to use deep learning (Hidden Attribute Models) for inferring personal attributes, such as profession, age, or family status, from conversations (Reddit discussions, movie scripts, and crowdsourced personal dialogues). 

The review presented by~\citet{Rastogi:2020:arXiv} covers personal health knowledge graphs (PHKG) for patients. PHKG is defined as a representation of aggregated multi-modal data including all the health-related personal data of a patient. It can be generated by inferring patient preferences over a given general-purpose KG (the authors mention the usage of entity linking).

The approach presented \tocheck{by}~\citet{Tigunova:2020:EMNLP} (built on top of work presented \tocheck{in~\citep{Tigunova:2019:WWW}}) proposes CHARM (Conversational Hidden Attribute Retrieval Model), a zero-shot learning method that creatively leverages keyword extraction and document retrieval in order to predict attribute values that were never seen during training. CHARM can be used for inferring attribute values in a zero-shot setting and extracting personal information from conversational utterances to populate the personal KB. The details of populating the personal KB with extracted attributes are not provided.

\citet{Ammar:2021:JMIR} study requirements and current shortcomings of electronic health records and use the Solid~\citep{Sambra:2016:techrep} platform to propose and prototype personal health libraries, and a mobile app using data from them. The personal health libraries are to be populated using existing Semantic Web technologies, such as Linked Open Data and the Web Annotation Data Model.\footnote{\url{https://www.w3.org/TR/annotation-model/}} The population process follows the REST principles. In addition, text summarization and knowledge mapping are mentioned as future directions to be explored in populating the PKG. 

\citet{Vannur:2021:PAKDD} discuss the subtasks of knowledge base population applied to extracting entities and relations from free text, specifically for the purpose of personal knowledge base population. It is defined as populating a knowledge base with personal information, albeit not necessarily limited to a PKG in the sense used in the present work. The focus of the paper is the subtask of extracting facts from free text. 

The personal health ontology used to generate PHKG~\citep{Gyrard:2018:ISWC} is presented by~\citet{Seneviratne:2021:arXiv} on the use case of diet recommendation. The authors use time series summarization to extract RDF triples from food logs that are then used to populate the PHKG. The theoretical considerations on modifications required in an existing framework to perform this, as well as a specific example are provided. However, implementational details are missing.

\citet{Chakraborty:2022:WWW} describe the use of PKGs to support academic researchers in their individual and collaborative research activities. The personal research knowledge graphs presented in the paper can be populated with the entities and relationships between them that are automatically extracted from different sources such as scholarly papers. Both manual management and automatic extraction from unstructured data using NLP techniques are mentioned.  As the authors follow the definition of \citet{Balog:2019:ICTIR}, entities present in public knowledge bases (e.g., Wikidata and Open Research Knowledge Graph\footnote{\url{https://www.orkg.org/orkg/}}~\citep{Jaradeh:2019:K-CAP}) are not integrated but linked.

\paragraph{Summary}
The majority of the papers discussed above provide broad and detailed descriptions of the PKG population aspect. Several works focus only on the problem of extracting information from existing data sources\tocheck{, e.g., \citep{Tigunova:2019:WWW, Tigunova:2020:EMNLP}} and adding them to the PKG\tocheck{, e.g., \citep{Gyrard:2018:ISWC}}. 
In these works, extraction is typically restricted to a selected set of predicates~\citep{Tigunova:2019:WWW, Tigunova:2020:EMNLP, Yen:2019:SIGIR}, or is not stated explicitly.
Interestingly, only two works mention the synchronization aspect\tocheck{~\citep{Groza:2007:ISemantics, Sambra:2016:techrep}}. These papers, inspired by Semantic Web technologies, propose standalone architectures for personal online datastores~\citep{Sambra:2016:techrep} or sharing data between users and applications~\citep{Groza:2007:ISemantics}.

\subsection{Representation and Management}
\label{sec:survey:represent_manage}

The way information is represented in a PKG, how data from external data sources are integrated with the ontology of the PKG, as well as the details of the concrete software used to manage the PKG, together comprise the second aspect of the PKG ecosystem. Here, the management system is considered without the 
 elements of administration interfaces which the management system nevertheless enables, which are discussed in \tocheck{Section}~\ref{sec:survey:utilization}. 

In the NEPOMUK project, \citet{Groza:2007:ISemantics} define a data resource as an RDF graph obeying some ontology or set of ontologies. To integrate external data, a Mapping Service translates RDF graphs from a source ontology to a target ontology. \citet{Groza:2007:ISemantics} also discuss access rights management and access control implemented with the NEPOMUK middleware, and mention the use of the Web Services Description Language (WSDL) to define services.\footnote{\url{https://www.w3.org/TR/wsdl/}} The work further envisions a shared information space based on peer-to-peer exchange, and consider both local and distributed storage of data.  

Solid~\citep{Sambra:2016:techrep} represents structured data as RDF, while unstructured data may be of any type, e.g., images, video, or free text. Application data is stored in documents with individual \tocheck{IRIs} for each resource. Solid specifies requirements for the personal online datastore (POD) management system and \citet{Sambra:2016:techrep} offer several Solid prototype servers to explore this specification. The management system must address RDF and non-RDF resource storage, basic data operations (Linked Data Platform\footnote{\url{https://www.w3.org/TR/ldp/}} operations and some Solid extensions of these), access control, and, optionally, complex data retrieval. Specifically, the users' data are stored in PODs and managed in a RESTful way, with SPARQL support for complex data retrieval operations.

\citet{Gyrard:2018:ISWC} describe the personal health knowledge graph (PHKG) in terms of component technologies, including the use of existing medical ontologies, but do not explicitly address the representation or management system of the PKG for health. They do discuss the integration of public ontologies and KGs from ontology catalogs such as BioPortal\footnote{\url{https://bioportal.bioontology.org/}} and Linked Open Vocabulary.\footnote{\url{https://lov.linkeddata.es/dataset/lov/}} They also mention some challenges in reusing existing ontologies, but do not share how these challenges are tackled in practice. 

\citet{Mazare:2018:EMNLP} present a large dataset of persona-based dialogues built using conversations extracted from Reddit, as well as end-to-end dialogue models trained on this dataset. The conversational dataset is used to extract a persona for some users, where a persona is represented as a set of sentences representing the personality of the responding user.  

\citet{Zhang:2018:ACL} present a dataset of chit-chats with personas and models trained on this dataset, conditioning next utterances on personas (for either or both sides of the dialogue) to be more engaging. Here the persona is represented as a set of maximum five sentences describing the persona in the first person. 

\citet{Lu:2019:NAACL} investigate different conversational agents to provide personalized customer service chat by exploiting customer profile information. The profile used for personalization is represented only by a few facts such as ``customer’s membership status, the order fulfillment method, the shipping carrier, whether the order is a single or multi-item order, and whether the order was eligible for cancellation at the time of contact.'' The profile information is not controlled by the person who is being profiled. 

\citet{Luo:2019:AAAI} first express the personal profile of the user as a concatenation of one-hot vectors, each representing a selected user attribute and its value. The user's profile model and preference model are then represented as neural embeddings to support the ranking of knowledge base items. 

\citet{Yen:2019:SIGIR} extract life events that are represented with quadruples of the form (object, predicate, subject, time). The quadruples are inserted into a knowledge base. The extraction of life events is integrated with predicates from the Chinese FrameNet ontology~\citep{Yang:2018:LREC}. 

\citet{Tigunova:2019:WWW} mention a personal KB without revealing any details on how it is constructed using the extracted attributes. However, the attributes are represented in the form of SPO triples. 

\citet{Rastogi:2020:arXiv} represent data using classes, entities, attributes, and relationships. They mention integration with public KGs and the use of predefined ontologies without giving any implementation details.

\citet{Tigunova:2020:EMNLP} represent extracted attributes in the form of attribute-value pairs. Here, the personal attributes of interest are only ``profession'' and ``hobby,'' and the possible values are drawn from the corresponding Wikipedia ``List of'' pages.\footnote{\url{https://en.wikipedia.org/wiki/List_of_professions} and \url{https://en.wikipedia.org/wiki/List_of_hobbies}.} 

\citet{Ammar:2021:JMIR} directly adopt the Solid approach and apply it to represent personal health data as various RDF and non-RDF data, distributed over potentially multiple PODs per patient/user, with a primary focus on the patient/user's RDF-based PKG of health information. 
They also mention knowledge mapping and the use of public knowledge bases and ontologies, but leave integration implementation details for future work. The management system is likewise directly derived from the Solid platform. 

\citet{Vannur:2021:PAKDD} are motivated by the problem of populating a PKG, but focus on extracting entities and relations. The exact representation of extracted information for PKG population is only discussed speculatively as the graph data could be exported to different formats, including RDF. The authors consider public KGs such as YAGO\footnote{\url{https://yago-knowledge.org/}} and the Person Ontology~\citep{Ganesan:2020:arXiv}. However, in practice they address integration by training models for entity classification and link prediction on existing datasets (OntoNotes\footnote{\url{https://catalog.ldc.upenn.edu/LDC2013T19}}~\citep{Hovy:2006:NAACL}, TACRED\footnote{\url{https://nlp.stanford.edu/projects/tacred/}}) with their respective attributes and entity types. 

\citet{Seneviratne:2021:arXiv} propose a new ontology for their PHKG, the Personal Health Ontology. We note that their ontology reuses concepts from existing ontologies such as the Statistics Ontology.\footnote{\url{https://stato-ontology.org/}} To the best of our knowledge the Personal Health Ontology is not publicly available. RDF triples are used to represent the data. 

\citet{Chakraborty:2022:WWW} suggest that a Personal Research Knowledge Graph (PRKG) may be modeled as a labeled property graph in Neo4j,\footnote{\url{https://neo4j.com/}} which may be serialized as an RDF graph for applications. Neo4j may be used without a predefined schema, and this is beneficial in the scenario of ongoing discovery. The facts in the PRKG are represented with SPO triples. The use of public knowledge graphs is mentioned, but integration such as ontology mapping is not explored. The management system is described only lightly, but some details may be implied by the choice of a Neo4j property graph implementation. \citet{Chakraborty:2022:WWW} do address access control, and find that role-based access control~\citep{Ferraiolo:2003:Book} on a node- or relation-level is not currently implemented in Neo4j, which they intend to correct. 

\paragraph{Summary}
The papers detailed above exemplify different approaches to the aspect of representation and management in the PKG ecosystem. 
The surveyed research is primarily concerned with extracting facts from unstructured data such as free text or chat dialogues. These are then to be inserted into a KG, where the final representation of each fact is expected to be SPO triples, and these facts are sometimes informed by pre-existing ontologies. This may be limited to conforming extracted predicates and entities to a pre-existing vocabulary. An interesting variation is given by \citet{Yen:2019:SIGIR}, who consider quadruples that extend the SPO format with an element of time. However, past research often describes the extraction of facts from unstructured data without explicitly addressing the format of the KG. In addition, some research on personalized KGs considers attribute-value pairs, which may imply that the person being profiled is the subject with extracted attributes and values as predicates and objects, respectively. Finally, \citet{Mazare:2018:EMNLP} and \citet{Zhang:2018:ACL} describe the use of full natural language sentences to personalize dialogue agents.
Broadly, the research surveyed here is primarily concerned with other aspects of the PKG ecosystem than representation and management, with implementation details of representation of knowledge, integration with ontologies, and the management software of the PKG largely relegated to either future work or defaulted to RDF format. \citet{Groza:2007:ISemantics}, \citet{Sambra:2016:techrep}, and \citet{Ammar:2021:JMIR} comprise exceptions to this rule, and present more complete systems in terms of our PKG ecosystem.

\subsection{Utilization}
\label{sec:survey:utilization}

Finally, we survey work in terms of the utilization of the PKG by its owner or by external services. More specifically, in our ecosystem, the owner of a PKG can administrate it (e.g., add new facts manually, grant access to external services or users), and external services can interact with the PKG depending on their access privileges.

In the platform proposed by \citet{Groza:2007:ISemantics}, the NEPOMUK API is intended to connect services to a semantic desktop that facilitates data sharing between the local user and other users and services. In particular, publish/subscribe services defined by SPARQL queries enable information streams to share updates with a community of users or applications. Furthermore, in the NEPOMUK platform, resources and the corresponding RDF descriptions can be added manually. The user also decides which information sources and users can be trusted. The access control system limits the individual user’s actions in the shared information space of the community. For example, a user can share a personal document only to a specific group of users.

Solid~\citep{Sambra:2016:techrep} specifies access control using WebID\footnote{\url{https://www.w3.org/2005/Incubator/webid/spec/ identity/}} to identify individuals and groups, and the WebAccessControl\footnote{\url{https://github.com/solid/web-access-control-spec}} ontology to describe access permissions to different resources for different WebIDs. The process to administrate the personal online data stores (PODs) is not described in detail, but would have to be implemented using the Linked Data Platform and Solid operations such as those used in updating or querying PODs. 
In Solid, the user may control applications' access to personal data, and a consistent API for the PODs supports interoperability and the users' freedom to switch between similar applications. 
\citet{Sambra:2016:techrep} present a number of typical applications, such as a contact manager, to show that Solid offers integration with multiple social Web applications for common day-to-day tasks.

\citet{Gyrard:2018:ISWC} argue that a personal health knowledge graph (PHKG) is well suited for a personalized health coach application. More particularly, they discuss the use-case of self-management of chronic diseases such as asthma. Later work by \citet{Rastogi:2020:arXiv}, \citet{Ammar:2021:JMIR}, and \citet{Seneviratne:2021:arXiv} also study use-cases related to the idea of a personalized health coach.
\citet{Rastogi:2020:arXiv} argue that PHKGs can be used to personalize recommendations from food platforms to encourage a healthy life style.
\citet{Ammar:2021:JMIR} exemplify the purpose of their proposed personal health libraries with a mobile app to support users' chronic disease self-management, and plan to expose the personal health libraries to third-party applications. Unlike the other work related to PHKGs in this survey, \citet{Ammar:2021:JMIR} discuss the administrate process by describing access control and the decoupling of data from applications accordingly.
\citet{Seneviratne:2021:arXiv} present the use-case of an application giving personal insight for Type 2 Diabetes self-management following clinical guidelines. They propose to make recommendations by reasoning over the PHKG containing clinical guidelines represented using OWL.

Several works~\citep{Zhang:2018:ACL,Lu:2019:NAACL,Luo:2019:AAAI,Mazare:2018:EMNLP,Tigunova:2019:WWW,Tigunova:2020:EMNLP} motivate the use of PKGs to create personalized dialogue systems, also referred as conversational agents. For example, \citet{Lu:2019:NAACL} propose to use profile information to adapt the behavior of a customer service conversational agent. In their work, \citet{Luo:2019:AAAI} describe a dialog system using a personalized profile where neither the extracted attributes and neural embeddings representing the user, nor the knowledge graph of items for the system to recommend are under the user's control.  
\citet{Tigunova:2019:WWW,Tigunova:2020:EMNLP} state that a personal knowledge base can be leveraged for personalization in downstream applications such as web-based chatbots and agents in online forums, however, additional details on ``how'' are not provided.

\citet{Yen:2019:SIGIR} mention that their personal knowledge base with life events can be used for memory recall and living support assistance. Specifically, they discuss the use of the personal knowledge base for question answering.

\citet{Gerritse:2020:ICTIR} study potential biases introduced by the use of PKG in conversational search. The authors discuss how and why the PKG can amplify biases in personalized services such as confirmation bias, e.g., when a user searches for a conspiracy theory, and data bias related to the population of the PKG.

\citet{Vannur:2021:PAKDD} assume a different scenario than our user-controlled PKGs scenario, but do motivate their work with the utilization of knowledge graphs containing personal information by enterprise application services to support data protection, fraud prevention, and business intelligence.

\citet{Chakraborty:2022:WWW} propose using personal research knowledge graphs (PRKG) for personalization in different scholarly applications like academic search engines and recommendation systems by sharing with them the user's personal data from the PRKG. Along the paper, the authors use the example of a researcher and how a personal assistant could benefit from the PRKG, as for the exploration a specific research space. 
With regards to administration, the owner has the possibility to add facts to the PRKG manually and can also check the ones added automatically. The user can also control how its personal data is shared. However, the authors do not discuss the interface between the owner and the PRKG. 

\paragraph{Summary}
Based on the papers discussed in this section, we observe that the administrative service allowing the user to directly interact with its PKG for access control or data update is largely disregarded. Indeed, only three papers provide details about it, two of which are from the field of Semantic Web technologies~\citep{Groza:2007:ISemantics,Sambra:2016:techrep}.
In terms of use, most of the selected papers focus on the creation of services exploiting the PKG to propose a personalized experience for each users. The application domain varies from health to scholarly work through customer service. Similarly, the type of services is diverse including conversational assistance and enterprise services. This illustrates the generic aspect of PKGs and their potential for many future applications, however, these should be aware of potential drawbacks such as biases~\citep{Gerritse:2020:ICTIR}.

\section{Challenges and Opportunities}
\label{sec:challenges}

The previous section has synthesized existing work on various aspects of PKGs. 
The main high-level observation that can be drawn from this survey is that few works focus on multiple aspects and none consider the PKG ecosystem as a whole.
Therefore, we start this section with a consideration of challenges and opportunities around the PKG ecosystem in Section~\ref{sec:challenges:ecosystem}, followed by the discussion of open issues around the individual aspects of population, representation and management, and utilization in Sections~\ref{sec:challenges:population}--\ref{sec:challenges:utilization}, respectively.
In order to illustrate our ideas in a more tangible and concrete manner, we will utilize the running examples introduced at the outset of the paper.
Specifically, Example~\ref{ex:training} with the personal trainer assistant will be used to elucidate near-term opportunities, while Example~\ref{ex:health} around the sharing of health information will aid in the illustration of longer-term challenges.

\subsection{Ecosystem}
\label{sec:challenges:ecosystem}

It is clear from the previous section that PKGs have been an active area of research, with new ground broken in various aspects. 
Some of these aspects are in isolation already well-established research areas, and some have been developed specifically for PKGs. However, these tasks remain underdeveloped from a holistic perspective in the context of the PKG ecosystem. 
Thus, a main overarching challenge remains: \emph{How do all the components within the PKG ecosystem fit together and how should they interact with each other?}

One of the main contributions of this work has been the treatment of the various aspects, components, and processes involved in the holistic realization of PKGs, and their organization in a unified architecture (cf. Fig.~\ref{fig:pkg-arch}).  We have also cast prior work within our framework, which illustrates its applicability.
At the same time, it is important to emphasize that it is an abstract, conceptual architecture.
Let us consider next what the creation of the Personal Trainer Application would entail in practice.

\begin{continuedexample}{ex:training}
    A straightforward solution would be to have a cloud-based back-end that stores the PKG and performs the various data operations (population, synchronization, etc.).  The mobile and web front-ends would fetch data from the back-end, display recommendations, send notifications, and allow the user to configure the service (features and integrations).  Effectively, the front-end serves as the administrative user interface, allowing the owner of the PKG to manage their history (i.e., provide full read and write access) and synchronize with specific external services (which have been integrated by the service provider via their respective APIs).
    The application may be designed to work only online (i.e., live internet connection is required) or could allow offline access to the recommendations (but not to the PKG itself).
\end{continuedexample}

While this example solution seems feasible, it focuses on a single application and relies on tailor-made components. 
Having a PKG for a single application defies its purpose and offers limited benefits beyond increased transparency on the service provider's end.
To unlock its potential, the PKG should amass data from multiple sources so that it could be utilized by multiple services. 
For example, there could come a time where data collected in the course of (uninjured) training using one application would be useful in the personalization of a rehab programme provided by a different application.
This requires standardization on the ecosystem level, that is, the use of shared vocabularies and communication protocols.
Having a shared data representation is critical to ensure that users remain free to share their own data to a different service provider (e.g., different fitness application). We could envision a ``PKG ready'' badge system, similar to an ISO classification, for applications/services that meet this established standard.
Note that this ``PKG readiness'' needs to extend to external services that allow for integration.
For example, while Facebook and YouTube allow users to download an archive of their activity on the platform, there is no programmatic access to the same data via an API.
This is not so much of a technical challenge, but rather a question of incentives.  Established social media platforms have a strong (and from a business perspective quite understandable) motivation to keep users ``locked in'' within their own ecosystem. 
There need to be either financial incentives or regulatory frameworks to convince service providers to open up access to data via APIs using the PKG standards (and thereby help promote the PKG approach to taking ownership of user data utilized for personalized services). 
We discuss more about representation and management below, under Section~\ref{sec:challenges:repr}.

In terms of the actual storage of PKGs, one possibility is to have them reside within accredited PKG hosting providers, similar to how other cloud-based services are hosted. This would provide users with a trusted and secure environment to store and manage their PKGs, following established standards for privacy and security. It would also allow users to take their PKG to a different hosting provider, should they decide to do so.
It is clear from Fig.~\ref{fig:pkg-arch} that the PKG and the management system around it are tightly coupled together, therefore the hosting provider would also need to provide an administration interface.  
Hosting providers could offer multiple options here, just like there are several options for virtual server management, including numerous open-source tools (cPanel, ISPManager, Webmin, etc.).  The user-friendliness of this administration interface is of critical importance, as ordinary users need to be able to manage large amounts of data through this interface; see Section~\ref{sec:challenges:utilization} below for further details.

It is worth pointing out that the PKG being under the owner's full control means that the owner can modify the PKG data. It is crucial that services take into account the possibility of incomplete or manipulated data when using PKGs. We illustrate this via our second example.

\begin{continuedexample}{ex:health}
    [Person] uses a PKG to manage their medical history, including records of treatments they have received in the past. They recently decided to share their PKG with a new doctor who would be treating them for a chronic condition. However, they were embarrassed about a particular medical procedure they had undergone in the past and did not want the new doctor to see it. Without much thought, they deleted the record of the procedure from their PKG.
    Unfortunately, the deleted record was relevant to the new doctor, who needed the information to properly diagnose and treat the patient's condition. The doctor's reliance on incomplete information from the PKG led to a misdiagnosis and ineffective treatment.
\end{continuedexample}

This example illustrates that the consequences could be severe and that ``personalized service'' must not be interpreted too liberally.  Receiving recommendations about what specialist to visit given certain symptoms may be a service, provided that the user understands that (1) the responsibility of providing truthful and complete input to the service via the PKG lies with them and (2) the recommendations are not to be taken as medical advice, but suggestions for consideration.
On the other hand, health care providers are responsible for educating patients about the importance of providing comprehensive and accurate medical history and relying on their own records, whenever possible.
A potential solution is presented in Section~\ref{sec:challenges:utilization}.

In summary, we identify the establishment of PKG standards and their adoption by service providers as the most important open challenges for the realization of a PKG ecosystem, both of which would require a combination of incentive structures and regulatory control.

\subsection{Population}
\label{sec:challenges:population}

Most research efforts relating to the aspect of PKG population concentrate on extracting structured data, like entities, relations, and attributes, from diverse data sources like tweets, academic articles, or food logs. However, the data extraction process usually restricts itself to a predetermined schema, e.g., a single vocabulary of predicates, which implies there is potential for further research. For example, how can a PKG population process discover novel predicates from unstructured data, assign these appropriately in the ontology, and subsequently apply these discovered predicates unambiguously? While data extraction from unstructured data is widely covered, only a few papers discuss the actual insertion of the extracted facts into the PKG, and even this coverage is quite light.

\begin{continuedexample}{ex:training}
The Personal Trainer Assistant presents an immediate opportunity to use data collected by wearable electronics for personal fitness. However, a choice of schema or ontology must be made to have a shared vocabulary for integrating facts from disparate data sources such as the user's personal free text notes and data collected by wearable electronics from different manufacturers, which may not be all based on the same schema. 
The choices on this level also need to be informed by the specific intended uses of the application. For example, recommending a training plan depends on an ontology for characterizing exercises and their nominal properties. Based on different user goals (e.g., improving running form versus lifting heavier weights), the fitness application would match intended effects of exercises with user goals. The PKG population process may then include user-specific considerations, as different fitness goals imply different priorities about which facts to record in the PKG in order to track progress. It is likely most users would not wish to see these nuances surfaced after choosing a particular fitness goal, but rather it should be handled unobtrusively by the Personal Trainer Assistant. 
Note also that, since diet is an important component of personal fitness, while manually recording consumed calories is a challenge, it would be very useful to extend the knowledge extraction for PKG population to include facts extracted from images, e.g., inferring nutritional composition facts from a photo of a meal, which can then be used to log the user's diet. 
\end{continuedexample}

In addition, the synchronization process has not been given enough attention, leaving room for future exploration on how to appropriately propagate changes made in the PKG to the private data sources. Synchronizing structured data with updates in the PKG is trivially automatable, but synchronizing may be more challenging in cases where the corresponding facts in private data sources are represented in unstructured form. Synchronizing unstructured data with updates in the PKG might need to rely on a notification system as proposed by \citet{Groza:2007:ISemantics} and \citet{Sambra:2016:techrep}.

For example, a record of past events should generally not be changed, but a current prose description should be kept up-to-date. In this case, the fact to update may have been detected and extracted from one place in free text, but may also be expressed in different places in the text. These other places expressing that fact would then also need to be updated to fully synchronize the private data source with the PKG update. Identifying and tracking all such locations in unstructured data may present a challenge. 

\begin{continuedexample}{ex:health}
Sharing Health Information depends in large part on extracting facts correctly and unambiguously from free text clinical notes. Besides the challenges of discovering and incorporating novel entities and predicates from free text into the PKG, an important unsolved challenge for sharing health information is synchronizing updates in the PKG back to the private data sources. In clinical practice, the current state of a patient's case often needs to be reflected in a technical description. 
The records that constitute a patient's history may be where a fact was extracted and used to populate the patient's PKG, and these should generally not be modified, since the sequence of changes in a patient's history may be clinically important information. However, if the reality of the patient's condition changes and the PKG is updated accordingly, then a text such as the patient profile description intended to be read first by new clinical practitioners joining the treatment team must be up-to-date and reflect current facts. For a patient undergoing multiple treatments concurrently, a single patient profile may not be appropriate, as a dentist may need to see a different summary than an endocrinologist. Thus, coordinating all the places in unstructured data where a factual update in the PKG should be synchronized may be a non-trivial challenge. 
\end{continuedexample}

From the two examples, we see challenges in both the process of populating a PKG with new extracted facts while conforming to a pre-determined target ontology, as well as the process of keeping selected external data sources synchronized with the current facts in the PKG. However, ubiquitous data capture through wearable electronics presents a great opportunity to apply the PKG concept in a useful and health-promoting manner.

\subsection{Representation and Management}
\label{sec:challenges:repr}

While knowledge representation and knowledge base management are well established research areas,
representation and management appears to be the least studied of the three identified aspects of PKGs.
What seems to be particular for the representation of PKGs over regular KGs is the need for detailed contextual descriptions of the PKG facts,
so that they can be correctly understood and interpreted at a later point in time and used for a wide range of purposes. 

\tocheck{
Another difference between PKGs and KGs is that while KGs are often centrally managed by a dedicated team, the responsibility of managing a PKG is ultimately its owner's, whose technical competence may be limited.
There is therefore a strong need for tools that allow owners to manage their PKG efficiently without caring about the intricacies of data management and knowledge representation.}

\begin{continuedexample}{ex:training}
    As the training preferences of the user are likely to be different depending on, e.g., the time of day, the time of the week, the time of year, and other personal events, such as traveling, the basic contextual data for all facts in the PKG should be recorded. 
    By analyzing the user's data and context, the personal trainer assistant is able to provide more effective recommendations. For instance, it can identify that while the user typically favors indoor training during the winter months, an exception to this pattern is outdoor jogging in light rain. 
    If training recommendations are given to the user based on these conclusions, it is important that an explanation for the recommendations is given which is rooted in the facts together with their context, so that the user can then better assess the given recommendations and adjust them as desired.
\end{continuedexample}

In this regard, a fruitful line of research would be to develop best practice modeling patterns for PKG statements that can gracefully cope with regular KG facts, provenance data, and statements about statements. This could take the form of a special purpose vocabulary designed to fit with a reification approach for statements, such as RDF's basic reification,\footnote{\label{foot:rdf}\url{http://www.w3.org/TR/rdf11-concepts/}} named graphs,\textsuperscript{\ref{foot:rdf}} or RDF*~\citep{Hartig:2017:ISWC}. A standardized vocabulary for PKG data would also be a huge benefit for the development of interoperable PKGs and PKG services. The PKG vocabulary and modeling patterns should accommodate expressing facts which may be stated using different, possibly incompatible, ontologies because the different facts in the PKG may be about unrelated domains. \tocheck{An early account of defining and using such a PKG vocabulary to translate natural language statements in a PKG via an easy-to-use web interface is demonstrated with promising results in recent work by \citet{bernard:2024:arXiv}.}

A greater challenge seems to be that traditional reasoning techniques may not be directly applicable in such a scenario.
On the one hand,
since a PKG will typically make use of different ontologies and contain complex facts, such as temporal facts and facts of different modalities,
we can assume that the required expressivity will go beyond the expressivity of OWL~2 DL.\footnote{\url{https://www.w3.org/TR/owl2-overview/}} OWL~2 DL is the most expressive fragment of OWL for which reasoning problems are decidable and reasoners are readily available, cf.~\citet{Horrocks:2006:KR}. As a consequence, standard OWL reasoning services may not be practically usable for most PKGs.
On the other hand,
the facts in a PKG may naturally be logically inconsistent, not because of modelling errors or poor data quality, but due to the the diversity and ``messiness'' of real-world facts. Since standard reasoning services do not cope well (or at all) with logical inconsistencies, other techniques are needed.

\begin{continuedexample}{ex:health}
    The medical history of the PKG owner could be inconsistent for a number of reasons. For example, it could be because different doctors have made conflicting assessments of the owner's condition, or because the owner has manipulated the contents of the PKG. In such cases, the inconsistency should be identified and attempted to be fixed if possible, while also not disrupting reasoning services that operate on other, unrelated parts of the PKG.
\end{continuedexample}

To allow for reasoning over (parts of) the PKG data, both for discovering new facts and for maintenance tasks such as consistency checks, new strategies to ensure that reasoning is possible must be developed.
One such strategy could be to carefully identify and isolate specific parts of the PKG for which reasoning may be performed using standard OWL reasoning techniques.
A different strategy is to develop more fault-tolerant reasoners capable of handling inconsistencies, cf.~\citet{Maier:2013:SWJ}.

\subsection{Utilization}
\label{sec:challenges:utilization}

The survey in Section~\ref{sec:survey} presents use cases for PKG\tocheck{s} in diverse domains, mainly for the creation of personalized services.
In our ecosystem, the same PKG is used by all the services, therefore, it represents resources with diverse formats (e.g., document, email, event, and text).
One challenge for the service providers is to use the PKG data in an efficient manner. This includes seamlessly supporting this diversity of resources in addition to anticipating that some information can be manipulated or missing for several reasons such as access restrictions.  

\begin{continuedexample}{ex:training}
	A PKG owner registers the following resources in their PKG: events attended, medical record documenting previous injuries, YouTube workout channels subscribed, and documents related to their diet.
	However, the owner decides to restrict access to its YouTube workout channels subscribed that can indicate how active the owner is and what type of exercise they prefer.
	Hence, the Personal Trainer Assistant should be able to suggest an adapted training plan despite the lack of information regarding the workout channels subscribed. 
	To overcome this absence of information, the Personal Trainer Assistant uses reasoning over the owner’s diet information to determine how active they are and uses this information as a substitute for the recommendation engine. Otherwise, the Personal Trainer Assistant can make a recommendation accompanied by a warning message indicating which missing information might improve the recommendation.
 \end{continuedexample}

One solution proposed in this example is that service providers could utilize the accessible PKG data 
by inferring new knowledge about a user with reasoning techniques. However, using reasoning might not be that trivial due to different challenges such as the ones presented in the previous section.
Therefore, the service providers could encourage the PKG owners to include truthful and complete information in their PKG and make them accessible via warning messages regarding missing information.

One long-term opportunity for PKGs lies in the administration process. Indeed, the survey in Section~\ref{sec:survey} shows that in general the owner of the PKG is not the user who receives personalized services, but the providers of these services. Consequently, the user does not have control over what is present in the PKG and who has access to it. 
Therefore, in the future, service providers ought to acknowledge that the ownership of the PKG belongs to the user and is operationalized via the administration process.
The administration process serves two main purposes: the direct update of data inside the PKG and the control of its access by its owner. 
As mentioned before, allowing the PKG owner to directly interact with PKG data comes with potential risks such as the addition of untruthful information (e.g., wrong date for the last radiation session) and the deletion of critical information (e.g., removal of an allergy). This implies that the administration process should provide some guarantees and safeguards regarding data manipulation. 
One solution may be for service providers to require certain permissions with respect to the user's PKG in order to provide the service, as is current practice in many mobile applications.
The access control needs to consider different types of entities, i.e., a user, a group of users, and a service, that need read and/or write privileges. 
The existing tools (e.g., WebID, WebAccessOntology, and JSON Web Tokens\footnote{\url{https://jwt.io/introduction}}) in the field of semantic technologies might be used to build such process.
Additionally, PKGs are not limited to a specific sample of the population, therefore, the administration process should be easy to use for everyone. An example would be to provide an intuitive user interface with menus and forms that abstracts the complex operations such as the creation and execution of SPARQL query to update the PKG. 
\tocheck{In recent work, \citet{bernard:2024:arXiv} present a solution for populating and querying a PKG using natural language statements, which are automatically translated to API calls (and ultimately to SPARQL queries) by a backend service.}

\begin{continuedexample}{ex:health}
    This example clearly illustrates the need for access control as it involves different actors (e.g., healthcare services and doctors). Indeed, one main question for the PKG owner is to determine what part of the PKG should be shared with whom. For example, they decide that the dentist does not need to know about the owner’s last eye test in order to treat them. Therefore, they regulate access to the PKG using a group of external users/services per medical specialty. To do so, the interface of the administration process has a feature to associate external users/services to a specific group and can select the type of access (read/write), e.g., with a radio button. 
    In terms of safeguards against data manipulation, medical service providers such as the dentist can require permission to keep their own unmodifiable copy of the PKG owner's medical history for the duration of the service provider relationship. Thus, if the PKG owner decides to remove critical information, e.g., removal of an allergy, from their PKG, the service providers can still make a reliable medical assessment.
\end{continuedexample}

To summarize, the administration process remains as a critical challenge for the adoption of PKG especially in the case of sensitive application domains where decisions can directly affect the PKG owner's life. The exploitation of PKG data also brings its share of questions to address, such as the treatment of missing or inaccessible data.

\section{Conclusion}
\label{sec:conclusion}

The aim of this study was to conduct a comprehensive survey and synthesis of the current research on personal knowledge graphs, with a focus on identifying critical gaps and key challenges that need to be addressed in order to enable the practical deployment of this technology.
An important discovery from this paper is that there is no clear consensus on the meaning of a PKG. To address this lack of clarity, we have made explicit and compared the different ways in which the term is used, and proposed a new definition that emphasizes (1) data ownership by a single individual and (2) the delivery of personalized services as the primary purpose.
Another key takeaway is the need to consider the larger ecosystem when conducting meaningful research on PKGs.
We have proposed an unifying architecture and identified three main aspects that are required for successful development and deployment of PKGs: (1) population, (2) representation and management, and (3) utilization.
Based on our survey of existing work, organized around these three key aspects, we make the following observations.
First, it is apparent that certain aspects of PKGs have received more attention than others, and that a comprehensive, holistic approach is lacking.
Second, different research communities (information retrieval, knowledge management, the Semantic Web, and natural language processing) can contribute to the development of PKGs in unique and complementary ways.
Third, while specific building blocks exist, integrating them into a practical solution, even if only a rudimentary one, remains a difficult task, made even more challenging by the need for service providers to support and promote this technology.  To drive adoption, service providers need to demonstrate the value that PKGs can bring to end-users.
Finally, it is important to recognize that the biggest obstacles involved in developing and deploying PKGs are not of a technical nature, but lie in broader societal and organizational issues.
To drive meaningful progress, it will be necessary for the research community, industry, and regulators to work together in a collaborative and strategic manner.

\bibliographystyle{elsarticle-harv} 
\bibliography{aiopen2022-pkg}

\end{document}